\documentclass[10pt,letterpaper]{article}
\usepackage[margin=1in]{geometry}
\usepackage{microtype}
\usepackage{graphicx}
\usepackage{subfigure}
\usepackage{booktabs} % for professional tables
\usepackage{amsfonts}
\usepackage{amssymb}
\usepackage{amsmath}
\usepackage{mathtools}
\usepackage{mathrsfs}
\usepackage{natbib}

\usepackage{hyperref}

\usepackage{authblk}

\usepackage{hyperref}
\usepackage{graphicx}
\usepackage{tabularx}

\author[1]{Ali Abbasi}
\author[2]{Parsa Nooralinejad}
\author[3]{Vladimir Braverman}
\author[2]{\\Hamed Pirsiavash}
\author[1]{Soheil Kolouri}
\affil[1]{\large Department of Computer Science, Vanderbilt University}
\affil[2]{\large Department of Computer Science, University of California, Davis}
\affil[3]{\large Department of Computer Science, Johns Hopkins University}

\title{Sparsity and Heterogeneous Dropout for Continual Learning in the Null Space of Neural Activations}

\begin{document}
\date{}

\maketitle
\begin{abstract}

Continual/lifelong learning from a non-stationary input data stream is a cornerstone of intelligence. Despite their phenomenal performance in a wide variety of applications, deep neural networks are prone to forgetting their previously learned information upon learning new ones. This phenomenon is called ``catastrophic forgetting'' and is deeply rooted in the stability-plasticity dilemma. Overcoming catastrophic forgetting in deep neural networks has become an active field of research in recent years. In particular, gradient projection-based methods have recently shown exceptional performance at overcoming catastrophic forgetting. This paper proposes two biologically-inspired mechanisms based on sparsity and heterogeneous dropout that significantly increase a continual learner's performance over a long sequence of tasks. Our proposed approach builds on the Gradient Projection Memory (GPM) framework. We leverage k-winner activations in each layer of a neural network to enforce layer-wise sparse activations for each task, together with a between-task heterogeneous dropout that encourages the network to use non-overlapping activation patterns between different tasks. In addition, we introduce two new benchmarks for continual learning under distributional shift, namely Continual Swiss Roll and ImageNet SuperDog-40. Lastly, we provide an in-depth analysis of our proposed method and demonstrate a significant performance boost on various benchmark continual learning problems. 

% Neural networks have been widely used to solve variety of classification and regression tasks. However, despite their state of the art performance in numerous applications, they suffer from catastrophic forgetting when put into a continual learning setup in which there are presented with multiple learning tasks in a sequential order. Many works in the literature tried to address the forgetting problem mostly at the expense of extensive memory usage by storing a considerable amount of data from the previous tasks. A promising direction, recently shown to mitigate catastrophic forgetting, is gradient projection which is mainly involved with taking a gradient step that reduces the loss function while being orthogonal to the subspace of the previous tasks. Orthogonality ensures that gradient moves along the null space of the previous tasks While gradient projection methods are effective in many continual learning experiments, their performance significantly degrades as the number of tasks increases. In this paper, we investigate the reason behind this reduction in the performance over long sequence of tasks. Furthermore, we propose a novel sparsity scheme at the logit level to alleviate catastrophic forgetting while using the capacity of our network to full extent.

\end{abstract}

\section{Introduction}

The capability to continually acquire and update representations from a non-stationary environment, referred to as continual/lifelong learning, is one of the main characteristics of intelligent biological systems. The embodiment of this capability in machines, and in particular in deep neural networks, has attracted much interests from the artificial intelligence (AI) and machine learning (ML) communities \citep{farquhar2018towards,von2019continual,parisi2019continual,mundt2020wholistic,delange2021continual}. Continual learning is a multi-faceted problem, and a continual/lifelong learner must be capable of: 1) acquiring new skills without compromising old ones, 2) rapid adaptation to changes, 3) applying previously learned knowledge to new tasks (forward transfer), and 4) applying newly learned knowledge to improve performance on old tasks (backward transfer), while conserving limited resources. Vanilla deep neural networks (DNNs) are particularly bad lifelong learners. DNNs are prone to forgetting their previously learned information upon learning new ones. This phenomenon is referred to as ``catastrophic forgetting'' and is deeply rooted in the \emph{stability-plasticity dilemma}. A large body of recent work focuses on overcoming catastrophic forgetting in DNNs. 

The existing approaches for overcoming catastrophic forgetting can be broadly organized into memory-based methods (including memory rehearsal/replay, generative replay, and gradient projection approaches) \citep{shin2017continual,farquhar2018towards,rolnick2019experience,rostami2020generative,farajtabar2020orthogonal},  regularization-based approaches that penalize changes to parameters important to past tasks \citep{kirkpatrick2017overcoming,zenke2017continual,aljundi2017memory,kolouri2020sliced,li2021lifelong, von2019continual}, and 3) architectural methods that rely on model expansion, parameter isolation, and masking \citep{schwarz2018progress,mallya2018packnet,mallya2018piggyback,wortsman2020supermasks}. Many recent and emerging approaches leverage a combination of the above mentioned mechanisms, e.g., \citep{van2019three}. Among the mentioned approaches, gradient-projection based methods \citep{farajtabar2020orthogonal,saha2020gradient,deng2021flattening,wang2021training,lin2022trgp} have recently shown exceptional performance at overcoming catastrophic forgetting. In this paper, we will focus on the Gradient Projection Memory (GPM) approach introduced by \cite{saha2020gradient}, and propose two biologically-inspired improvements for GPMs, which are algorithmically simple yet they lead to a significant improvement in performance of the GPM algorithm.  

Sparsity is a critical concept that commonly occurs in biological neural networks. Sparsity in biological systems emerges as a mechanism for conserving energy and avoiding neural activations' cost to an extent that is possible. Neuroscientists estimate only 5-10\% of the neurons in a mammalian brain to be active concurrently \citep{lennie2003cost}, leading to sparse decorrelated patterns of neural activations in the brain \citep{yu2014sparse}. Similarly in ML, sparsity plays a critical role for obtaining models that are more generalizable \cite{makhzani2013k} and are robust to noisy inputs and adversarial attacks \cite{guo2018sparse}. Moreover, sparse neural networks are critical in deploying power-efficient ML models.  In continual learning, various recent studies show the role of sparsity in neural parameters \citep{schwarz2021powerpropagation} or in neural activations (i.e., representations) \citep{aljundi2018selfless,jung2020continual,gong2022learning} in overcoming catastrophic forgetting.  The existing work combine sparse model with regularization based continual learning or memory replay approaches and show significant improvements over baseline performances. These methods enforce sparsity via inhibition of activations \citep{aljundi2018selfless,ahmad2019can}, through regularization \citep{jung2020continual}, or by optimization reparameterization during training \citep{schwarz2021powerpropagation}. In this paper, we consider continual learning under power constraints and study the role of sparsity for continual learning in the nullspace of neural activations. We propose an adaptation of the k-winner algorithm \citep{ahmad2019can}, to enforce sparsity and encourage decorrelated neural activations among tasks. The power consumption of our proposed algorithm at the inference is significantly smaller than that of the GPM, while it provides a demonstrable advantage over GPM in learning a large sequence of tasks. 

{\bf Contributions.} Our main contributions in this paper are: 1) proposing a modification of the k-winner algorithm that leads to a significant gain in continual learning from long sequences of tasks, 2) showing that sparse activations through the modified k-winner algorithm result in low-rank neural activation subspaces at different layers of a neural network, 3) introducing conditional dropout as a mechanism to encourage decorrelated activations between different tasks, 4) introducing \emph{Continual Swiss Roll} as a lightweight and interpretable, yet challenging, synthetic benchmark for continual learning, and 5) introducing the ImageNet SuperDog-40 as a `distributional shift' benchmark for continual learning. We demonstrate the effectiveness of our proposed method on various benchmark continual learning tasks and compare with existing approaches.  

\section{Related Work}
{\bf Gradient-projection based approaches.} These methods rely on the observation that during learning a new task the Stochastic Gradient Descent (SGD)- or its variants- is oblivious to the past knowledge \cite{farajtabar2020orthogonal}. Hence, generally speaking, these methods rely on carrying the gradient information from previous tasks, whether by storing samples from old tasks or by learning gradient subspaces, and adjust the gradient for the current task such that it does not unlearn or interfere with previous tasks. In Orthogonal Gradient Descent (OGD), \cite{farajtabar2020orthogonal} store a memory of the loss gradients on previous tasks and project the loss gradient for the current task onto the nullspace of the previous gradients in the memory. This method requires a growing memory in the number of tasks. This idea is extended in \cite{saha2020gradient} and \cite{wang2021training} take this idea one step further and propose to carry and update a gradient subspace for previous tasks, and project the current loss gradient onto the nullspace of previous gradients, requiring a fix-size memory. This core idea has been extended in numerous directions in recent publications, e.g.,  \cite{deng2021flattening}. One major shortcoming of the gradient projection approaches, however, is that, by design, these methods provide no backward transfer. This shortcoming has led to more recent studies \citep{lin2022trgp}, that relax the orthogonal projection and allow for gradients to live in the subspace of `similar' previous tasks leading to some backward transfer among tasks.

{\bf Neural networks with low-rank activation subspaces.} low-rank structure of neural activations (i.e., low-rank representations) is a relatively understudied research area. The existing work, for instance, show that inducing low-rank structure on neural activations could result in networks that are more robust to adversarial attacks, and are also, not surprisingly, more compressible \cite{sanyal2019intriguing}. As another example, \cite{chen2018subspace} show the effectiveness of a low-rank constraint on the embedding space of an autoencoder in clustering applications. More relevant to our work, is the fantastic work by \cite{chaudhry2020continual}, in which the authors learn tasks in orthogonal low-rank vector subspaces (layer-wise) to minimize interference between tasks. %To enforce the loss gradients to be orthogonal to one another, they use similar ideas to that of the orthogonal gradient projections. 
We approach the problem from a different angle. Our rationale is that having a lower rank subspace for neural activations of previous tasks, would lead to a larger nullspace for projecting the loss gradient of the current task. In other words, enforcing lower rank neural activation subspaces for previous tasks leads to learning new tasks with a better accuracy as the projection of the loss gradient onto the nullspace introduces a smaller error. Finally, we emphasize that having low rank subspaces for layer-wise neural activations does not imply having sparse activations, and hence it does not provide a lower power consumption. 

{\bf Sparse neural activations.} Sparsifying neural networks has been an active field of research with a focal point of model compression and minimizing power consumption. Sparsity in neural networks could refer to sparsity in parameters (i.e., synapses), or sparsity in neural activations (i.e., sparse representations). In Powerpropagation, \cite{schwarz2021powerpropagation} propose a reparameterization scheme that encourages larger parameters to become larger and the smaller ones become smaller during training. The authors then showed that pruning a network that is trained with powerpropagation would lead to a very sparsely connected network, that can be combined with the PackNet algorithm \cite{mallya2018packnet} and effectively solve continual learning problems on various benchmarks for large sequences of tasks. Similar to PackNet \cite{mallya2018packnet}, the resulting approach in \citep{schwarz2021powerpropagation} requires task identities during the inference/test time. In Selfless Sequential Learning, \cite{aljundi2018selfless} show the role of sparse neural activations for regularization-based approaches in continual learning, e.g., \cite{kirkpatrick2017overcoming,aljundi2017memory}. In this work, we leverage k-winner activations \citep{ahmad2019can} to induce sparsity in our network and show the benefit of this induced sparsity in the GPM framework. 

{\bf Nonoverlapping activations.} Using non-overlapping neural representations for overcoming catastrophic forgetting is not a new idea \cite{french1991using,french1999catastrophic}. But this idea has recently revived in the continual learning community \citep{masse2018alleviating,mallya2018packnet,chaudhry2020continual}. \cite{mirzadeh2020dropout} makes an interesting observation that the vanilla Dropout  helps with the catastrophic forgetting problem via inducing an implicit gating mechanism that promotes non-overlapping neural activations. In our work, we introduce a task conditional Dropout that encourages non-overlapping sparse activations between tasks. We show that the addition of task-conditional Dropout to our sparse neural activation framework provides an additional boost in the performance of GPM. 

\section{Method}

\subsection{Problem Formulation}
Following the recent works in continual learning \citep{saha2020gradient,lin2022trgp}, we consider the setting where the learning agent learns a sequence of tasks $\mathbb{T}=\{t\}_{t=1}^T$ that arrive sequentially. Each task $t$ has its corresponding data set $\mathbb{D}_t=\{(x_{t,i}\in\mathbb{R}^d, y_{t,i}\in\mathbb{R}^k)\}_{i=1}^{N_t}$, where $x_{t,i}$ is the d-dimensional sample, and $y_{t,i}$ is its corresponding label vector. Also, while we consider the tasks to be supervised learning tasks, our approach is applicable to unsupervised/self-supervised tasks as well. We consider a neural network with $L$ layers, where each layer's weights are denoted by $W^l$ leading to the parameter set $\mathbb{W}=\{W^l\}_{l=1}^L$. For the $i$'th sample from the $t$'th task, we denote the neural activations at the $l$'th layer of the network as $x^l_{t,i}$ with $x^1_{t,i}=x_{t,i}$, and $x^{l+1}_{t,i}=f(W^l x^l_{t,i})$ for $\forall l \leq L$ where $f$ is the operation of the network layer.  Lastly, we assume that the agent does not have access to data from previous tasks while learning a new task (i.e., no memory buffer), and that the agent does not have knowledge about task IDs during training or testing. Note that the model still needs to know about the task boundaries during training but does not require the task IDs.

The core idea behind the gradient projection approaches \citep{farajtabar2020orthogonal,saha2020gradient,deng2021flattening,wang2021training} is to update the model parameters on the new task, such that it guarantees preservation of neural activations on previous tasks. Of particular interest to us is the Gradient Projection Memory (GPM) framework of \cite{saha2020gradient}, which we briefly describe here. Let $S_t^l=span(\{x_{t,i}^l\})$ denote the subspace spanned by the neural activations of the $t$'th task at layer $l$. Then while learning task $\tau$, GPM enforces the gradient updates for the $l$'th layer of the network to be in the null-space of $S_t^l$ for $\forall t< \tau$ by projecting the gradients onto the null space. Let $\mathbb{W}_\tau=\{W_{\tau}^l\}_{l=1}^L$ denote the model parameters after learning task $\tau$. Then under the gradient projection constraint, we can see that, after learning task $\tau$, the neural activations for task $t$ remain unchanged:
\begin{align}
    W_\tau^l x^l_{t,i}=(W_t^l+\Delta W_{\tau,t}^l) x^l_{t,i}= W_t^lx^l_{t,i}+\Delta W_{\tau,t}^lx^l_{t,i} =  W_t^lx^l_{t,i}
    \label{eqref:gpm}
\end{align}
where $\Delta W_{\tau,t}^l=W^l_\tau-W^l_t$ and $\Delta W_{\tau,t}^lx^l_{t,i}=0$ follows from the fact that we only optimize the network in the null spaces of  $S_t^l$s. GPM and other gradient-projection based approaches have shown remarkable performance on overcoming catastrophic forgetting.
However, despite their great success, one can observe two drawbacks with such gradient-projection approaches: 1) the framework does not allow for backward transfer, and 2) the network can saturate very fast, i.e., the null space of $S_t^l$ could become empty after a few tasks leading to intransigence. Regarding the first drawback, the lack of backward transfer have led to follow-up work \citep{lin2022trgp} that relax the orthogonal projections according to a notion of task similarities in favor of backward transfer. As for the second drawback, which is the focal point of our work, \cite{saha2020gradient} approximate the null-space by discarding dimensions with low variance of activation (i.e., small eigenvalues). While this practical strategy ensures avoiding intransigence it can violate \eqref{eqref:gpm} and lead to catastrophic forgetting of the old tasks.  We observe that if $S_t^l$s are low-rank subspaces, then the null space remain to be large and the network can be trained on more tasks. In this case, one can use a regularization to enforce low-rank activations, for instance through the nuclear norm of the covariance of neural activations. The low-rank constraint, however, does not enforce sparsity in the activations, which can help in reducing the power consumption at the inference time. Next, we propose our neural network with sparse activations for low-power continual learning.

\subsection{$k$-Winner Sparsity}

\begin{figure}[t!]
    \centering
    \includegraphics[width=.75\linewidth]{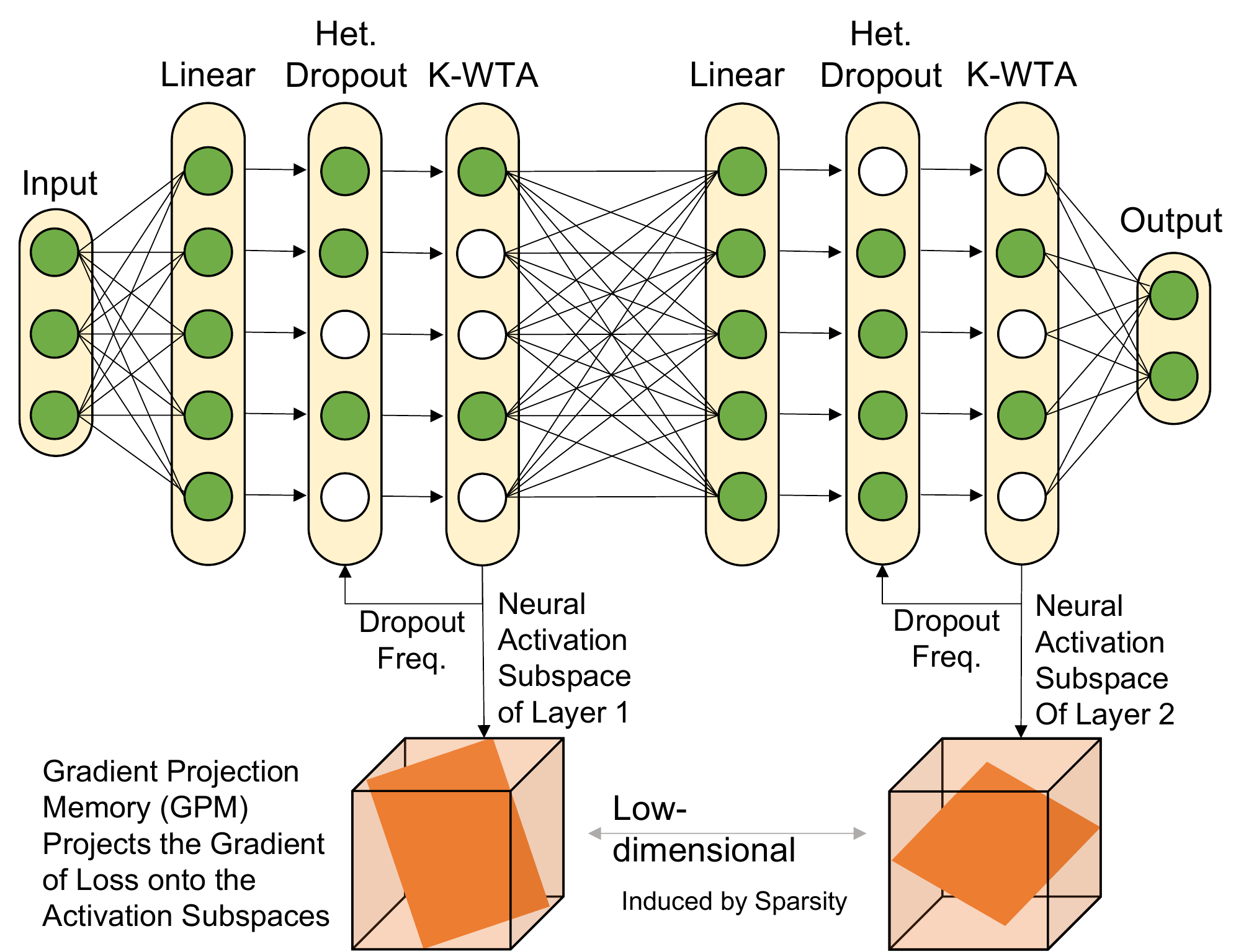}
    \caption{Depicting the overall structure of a multi layer perceptron along-with the k-winner activation and the proposed heterogeneous dropout layers.}
    \label{fig:structure}
\end{figure}

We leverage sparsity in neural activations with the target of: 1) reducing power consumption, and 2) reducing the rank of the subspaces $S_t^l$ for $\forall t\in\mathbb{T}$ and $l\in\{2,...,L\}$. Of course, sparsity of $x^l_{t,i}$ does not guarantee a low rank $S_t^l$, e.g., even one-sparse activations could lead to neural activation subspaces that are full rank. However, we show numerically that neural networks trained with sparse activations often form low-dimensional activation subspace, $S_t^l$. 

{\bf Sparse Activations:} Following the recent work of \cite{ahmad2019can} we leverage k-winner activations to induce sparse neural representations. The framework is similar to the work of \cite{majani1988k}, \cite{makhzani2013k}, and \cite44{Srivastavacompete2compute}. In short, each layer of our network follows, $x_{t,i}^{l+1}=f(W^lx_{t,i}^{l})$ where $f(\cdot)$ is an adaptive threshold corresponding to the $k$'th la rgest activation. Hence, only the top-$k$ activations in each layer are allowed to propagate to the next layer leading to $\|x_{t,i}^{l+1}\|_0\leq k$. One advantage of the k-winner framework is that we have control over the sparsity of neural activations through parameter $k$. 

But why would training with the k-winner activations lead to a low-dimensional activation subspace, $S_t^l$, at the $l$'th layer for task $t$? The answer is implicitly presented in the work of \cite{ahmad2019can}, where the authors observe that using k-winner activations in a network would lead to a small number of neurons that dominate the neural representation, i.e., they become active for a large percentage of input samples. This observation is also aligned with the previous observations made by \cite{makhzani2013k} and \cite{cui2017htm}.   \cite{ahmad2019can} view the dominance of a few neurons as a practical issue and solve this issue through a novel boosting mechanism that prioritizes the neurons with a lower frequency of activations. In our framework, however, we prefer to have a small set of dominantly active neurons in each layer of the network for each task, which translates to having a low-dimensional activation subspace, $S_t^l$. Hence, unlike the work of \cite{ahmad2019can}, we do not require any boosting within a task. 

Not having boosting while learning a task, could lead to dominant neurons across tasks, which could significantly reduce the overall performance of the model. To address this newly emerged practical issue in continual learning, and motivated by the approaches using non-overlapping neural representation for continual learning, we propose a conditional dropout between tasks, that encourages diverse neural activations between different tasks.

\subsection{Heterogeneous Dropout}
\label{subsec:dropout}

While training a network on a task, we keep track of the frequency of the neural activations. In short, we assign an activation counter per neuron, which increments when a neuron's activation is in the top-k activations in its layer (i.e., the neuron is activated). Let $[b_{t}^l]_j$ denote the activation counter for the $j$'th neuron in the $l$'th layer of the network, after learning task $t$. Note that $[b_{t}^l]_j$ represents the number of times the j'th neuron in layer $l$ was in the top $k$ activations over all previously seen tasks $\tau\in[1,...,t]$. Then, while learning task $t+1$, we would like to encourage the network to utilize the less activated neurons. To that end, we propose a dropout \citep{srivastava2014dropout} mechanism that favors to retain neurons that are less activated in previous tasks. We define a binary Bernoulli random variable, $[\delta_{t+1}^l]_j$, for the $j$'th neuron in layer $l$ during training on task $t+1$ that indicates whether the neuron is disabled by the dropout or not. In particular, we set $P([\delta_{t+1}^l]_j=1)=[p_{t+1}^l]_j$ for:
\begin{align}
    [p_{t+1}^l]_j=exp(-\frac{[b_{t}^l]_j}{\max_j [b_{t}^l]_j} \alpha)
    \label{eq:dropout}
\end{align}
where $\alpha>0$ is a hyper-parameter of our proposed dropout mechanism. Larger $P$ corresponds to less dropout and larger values of $\alpha$ lead to a more stringent enforcement of non-overlapping representations. %In our experiments, we choose $\alpha\propto\frac{1}{k}$ so that when $k$ is large we do not dropout a large portion of the network. \hamed{COMMENT: this is not clear why we do this}

We call the proposed dropout a heterogeneous dropout as the probability of dropout is different for various neurons in the network. Importantly, the probability of dropout is directly correlated with the frequency of activations of a neuron for previous tasks.  Hence, heterogeneous dropout will encourage the network to use non-overlapping neural activations for different tasks. Interestingly, the proposed heterogeneous dropout induces a ``lifetime sparsity'' of a neuron, which is well studied in the neuroscience literature \citep{beyeler20163d}.

In the following section, we first show that the k-winner framework leads to low-dimensional neural activation subspaces, $S_t^l$. We show that this low-dimensional structure enables learning more tasks with less forgetting using the GPM framework, leading to a significant performance boost. Finally, we show that our heterogeneous dropout encourages non-overlapping neural activations, which provide an additional boost in the performance of GPM over a large sequence of tasks. 

\section{Numerical Experiments}

\begin{figure}[t!]
    \centering
    \includegraphics[width=\linewidth]{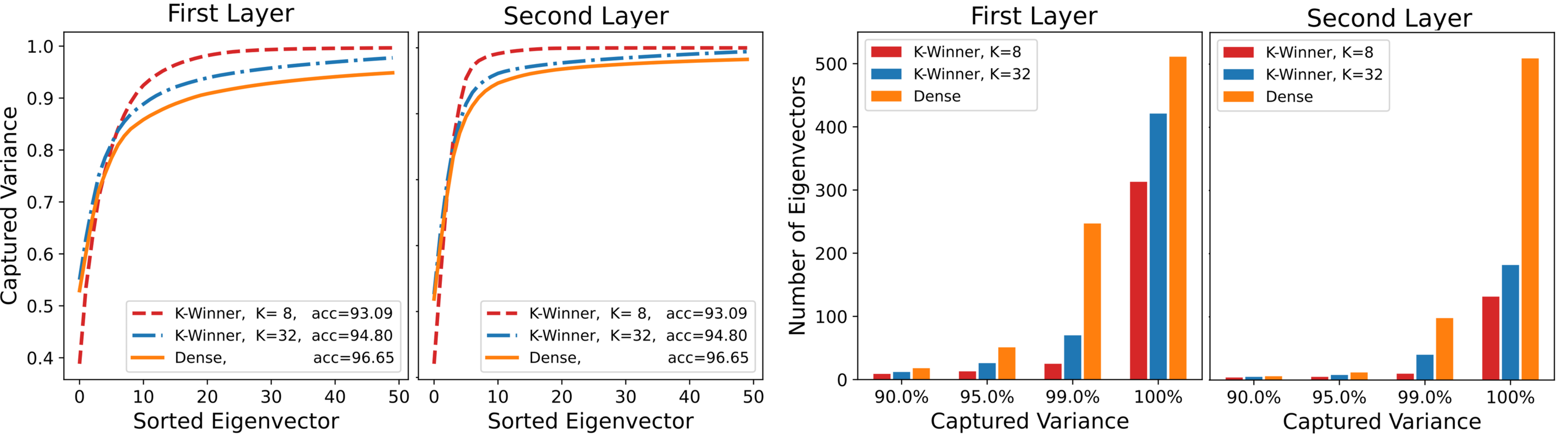}
    \caption{Demonstrating that a multi-layer perceptron, with two hidden layers of size $512$ and with k-winner activations exhibits low-dimensional neural activation subspaces, $S_t^l$, in its layers. The left column shows the captured variance as a function of top 50 eigenvalues/eigenvectors of the activations' covariance matrix for the densely trained model (no sparsity), and the networks trained with k-winner activations for $k=8$ and $32$, in the two layers of the networks. The right column shows the approximate dimensionality of $S_t^l$ at different thresholds on captured variance, $\epsilon_{th}$, for the two layers. It can be seen that networks with k-winner activations consistently provide lower-dimensional activation subspaces (at all thresholds). The reported Accuracy is the average accuracy over all tasks at the end of the training for the last task.}
    \label{fig:capturedVariance}
\end{figure}

\subsection{Training with k-winner Activations Leads to Low-Dimensional Subspaces}
\label{sec:lowDim}

 Our rationale is that having low-dimensional activation subspaces, $S_t^l$, while using gradient-projection based continual learning algorithms, like GPM, would lead to less gradient projection error in learning subsequent tasks (i.e., we have a larger null-space). On the other hand, we prefer sparsely activated networks to reduce the power consumption in a continual learner. However, a network with sparse neural activations does not necessarily guarantee low-dimensional activation subspaces. In this section, as one of our core observations, we show that using the k-winner activations to learn a task leads to low-dimensional activation subspaces, $S_t^l$, at each layer of the network. Then, in the subsequent sections, we show that these low-dimensional activation subspaces lead to a significantly better learning performance on a long sequence of tasks. 

We start by training a multi-layer perceptron with two hidden layers of size $512$ on the MNIST dataset \citep{lecun1998gradient}. We first train the model without the k-winner activations, i.e., with $k=512$, and call this model the \emph{Dense} model. Next, we train the model using k-winner activations with $k=8$ and $k=32$. For each model we calculate the activations $x_i^l$ for all $i$ and for $l\in\{1,2\}$. We then calculate the Singular Value Decomposition (SVD) of $X^l=[x_1^l,...,x_{N}^l]$,  calculate the eigenvalues of the covariance matrix (i.e., squared singular values), and sort the eigenvectors according to their eigenvalues. Lastly, we calculate the percentage captured variance (i.e., cumulative sum of sorted eigenvalues devided by the sum of the eigenvalues) for these networks. Figure \ref{fig:capturedVariance} visualizes the captured variance as a function of the first $50$ eigenvectors for both layers of the network for all three models. We observe that the models trained with k-winner activations lead to lower dimensional activation subspaces, $S_t^l$. Importantly, we note that \cite{saha2020gradient}, and subsequent works, do not calculate the exact null-space of the activation subspaces, but they approximate the null-space by zeroing out the eigenvalues that capture a small percentage of the variance. This is done via thresholding the captured variance at $0<\epsilon_{th}\leq 1$. Note that the exact null-space may be very small due to very small but non-zero eignevalues. To that end, we show that for various thresholds of the captured variances, i.e., $\epsilon_{th}\in\{0.90,0.95,0.99,1.00\}$, the networks using k-winner activations require fewer eigenvectors. Finally, the low-dimensional activation subspaces are achieved without a major loss in the accuracy of the trained networks.

\subsection{Heterogeneous Dropout for non-overlapping representations}

\begin{figure}[t!]
    \centering
    \includegraphics[width=0.95\linewidth]{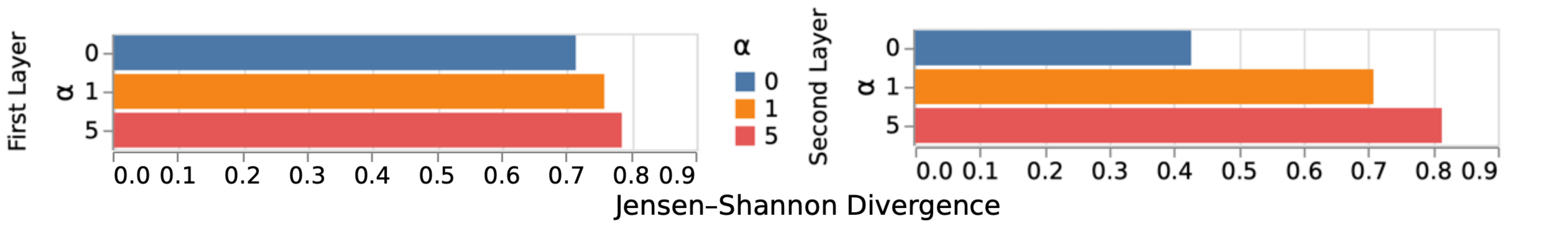}
    \caption{The Jensen-Shannon divergence (JSD in \eqref{eq:jsd}) between activations' probability mass functions, $q^l_1$ and $q^l_2$, for $l\in\{1,2\}$. The plot quantifies the overlap between neural activations of two tasks as a function of our dropout hyperparameter $\alpha$ (See  \eqref{eq:dropout}). Higher JSD values mean lower overlap. We can see that we are able to decrease the overlap between neural activations by increasing $\alpha$ ($\alpha=0$ results in no dropout).}
    \label{fig:overlap}
\end{figure}

Here we numerically confirm that our heterogeneous dropout leads to fewer overlaps between neural representations of different tasks. For these experiments, we use the GPM algorithm on a model with k-winner activations and learn two tasks from Permuted-MNIST sequentially, where the first task is MNIST and the second task is a permuted version. After training on Task 1, we calculate the number of times each neuron is activated for all samples in the validation set of Task 1. Then, we learn Task 2 using gradient-projection and afterwards calculate the number of times each neuron is activated for all samples in the validation set of Task 2. For task $t$ and for the $j$'th neuron in layer $l$, we denote the neural activations on the validation set as $[\nu^l_t]_j$. Note that $\nu^l_t$ is different from $b_t^l$ introduced in Subsection \ref{subsec:dropout}, as it is calculated on the validation set (as opposed to the training set), and it is calculated per task, while $b_t^l$ is the accumulation of activations over all tasks. Let $\bar{\nu}_t^l=\sum_{j=1}^{d_l} [\nu_t^l]_j$ where $d_l$ is the number of neurons in the $l$'th layer, then we can define a probability mass function of activations for each layer as $[q^l_t]_j=[\nu^l_t]_j/\bar{\nu}^l_t$. Finally, we measure the neural activation overlap between tasks $t_1$ and $t_2$, via the Jensen-Shannon divergence (i.e., the symmetric KL-divergence) between their neural activation probability mass functions, i.e,: 
\begin{align}
JSD(q^l_{t_1}||q^l_{t_2})= \frac{1}{2}\sum_j q^l_{t_1} log (\frac{2q^l_{t_1}}{q^l_{t_1}+q^l_{t_2}})+\frac{1}{2}\sum_j q^l_{t_2} log (\frac{2q^l_{t_2}}{q^l_{t_1}+q^l_{t_2}})
\label{eq:jsd}
\end{align}

Figure \ref{fig:overlap} measures the overlap between neural representations (between Task 1, MNIST, and Task 2, a Permuted MNIST) when the networks are trained with and without our heterogeneous dropout and for different values of $\alpha$. Higher Jensen-Shannon divergence means less overlap. In short, $\alpha=0$ means no dropout, and we confirm that higher $\alpha$ translates to less overlap (higher JS-divergence) between the neural representations.  Finally, we note that the choice of $\alpha$ truly depends on the amount of forward transfer we expect to see between two tasks. Lower values of $\alpha$ are beneficial when we expect our network to rely more on previously learned features (i.e., forward transfer from previous tasks), while higher values are preferred to learn new features for the new task.

In what follows, we combine k-winner sparse activations with our heterogeneous dropout, utilize Gradient Projection Memory (GPM) as our core continual learning algorithm, and demonstrate the benefits of sparsity and non-overlapping representations in learning long sequences of tasks in GPM. 

\subsection{Continual Swiss Roll}

% \begin{figure}[t!]
%     \centering
%     \includegraphics[width=.8\linewidth]{plots/ContinualSwissRoll.pdf}
%     \caption{The Continual Swiss Roll dataset consisting of 50 binary classification tasks (a), GPM's result (b), and our results (c) after sequentially training on all tasks. For both methods, we use a single head multi-layer perceptron (MLP) with two hidden layers of size 1024. We selected $k=64$ for this experiment.}
%     \label{fig:swiss}
% \end{figure}

\begin{figure}
\centering
\begin{tabular}{cccc}
    \includegraphics[width=.22\linewidth]{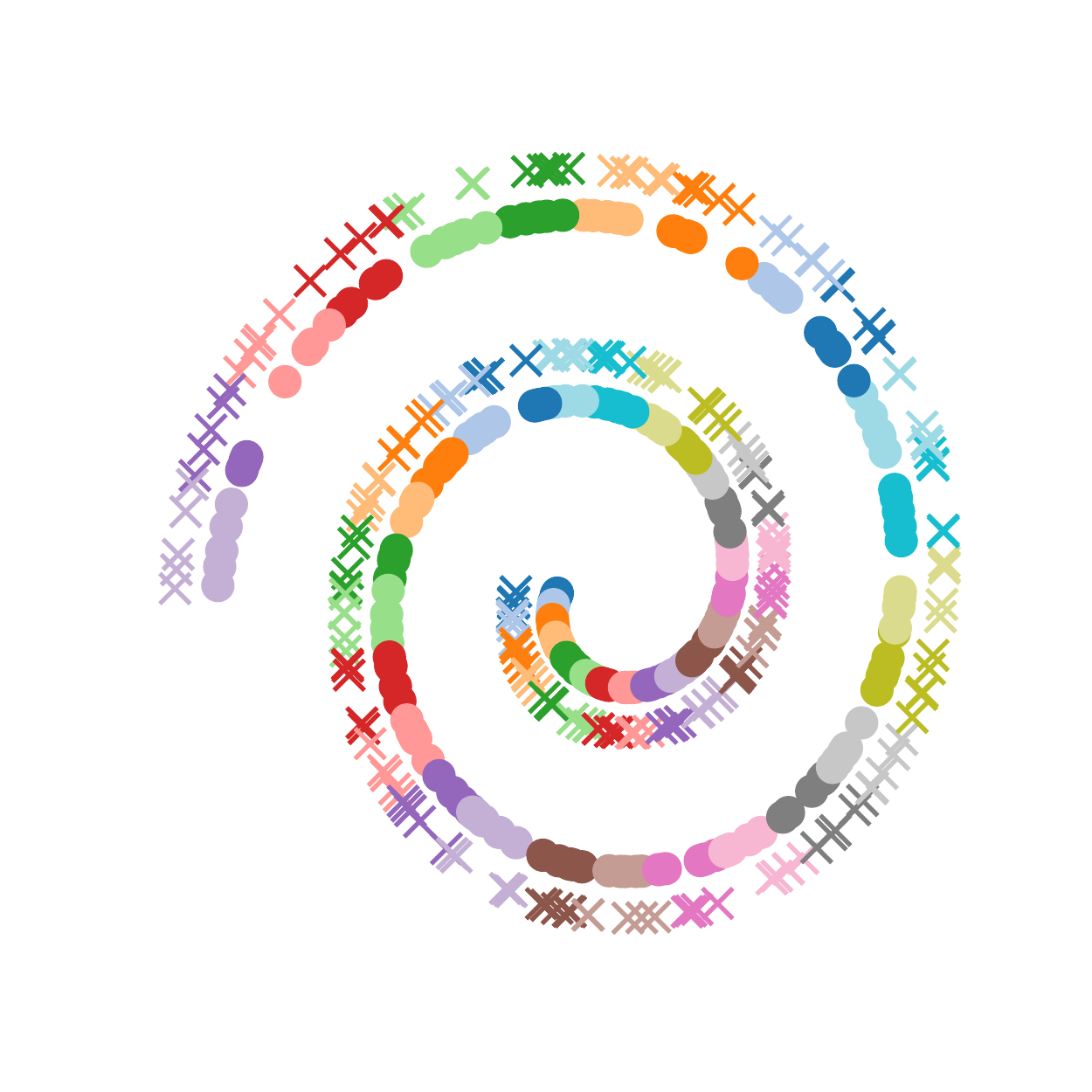} &  
  \includegraphics[width=.22\linewidth]{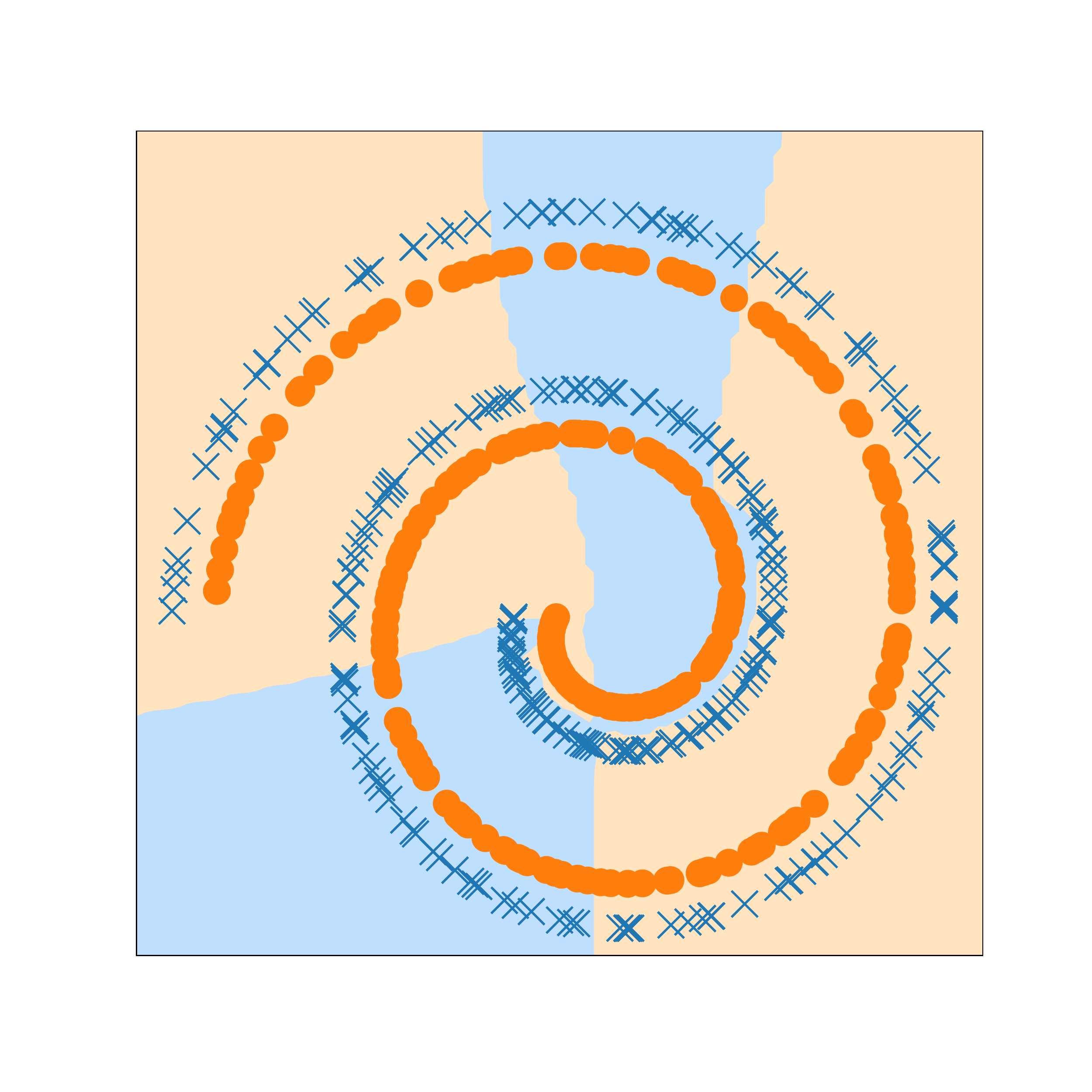} &   \includegraphics[width=.22\linewidth]{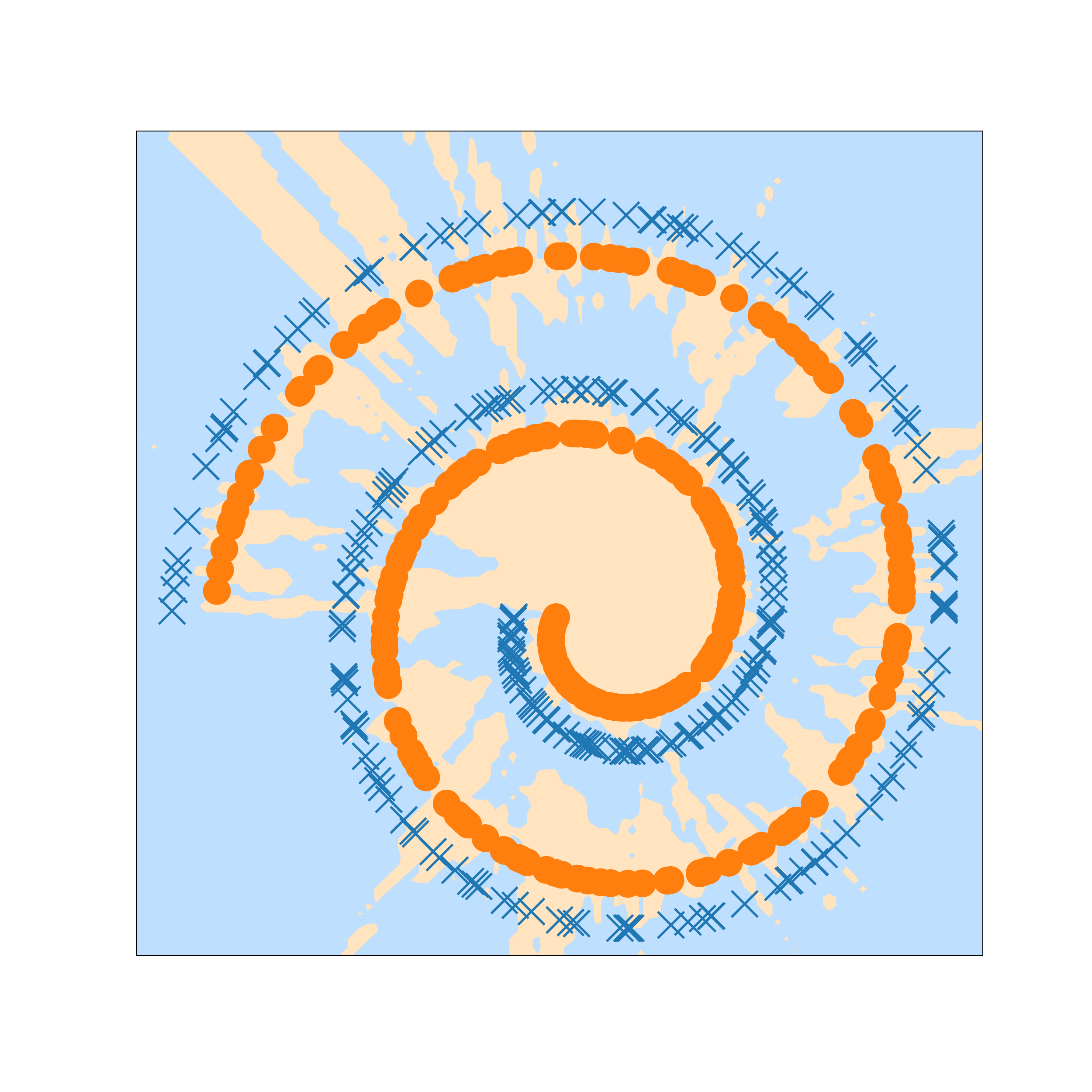} &
  \includegraphics[width=.22\linewidth]{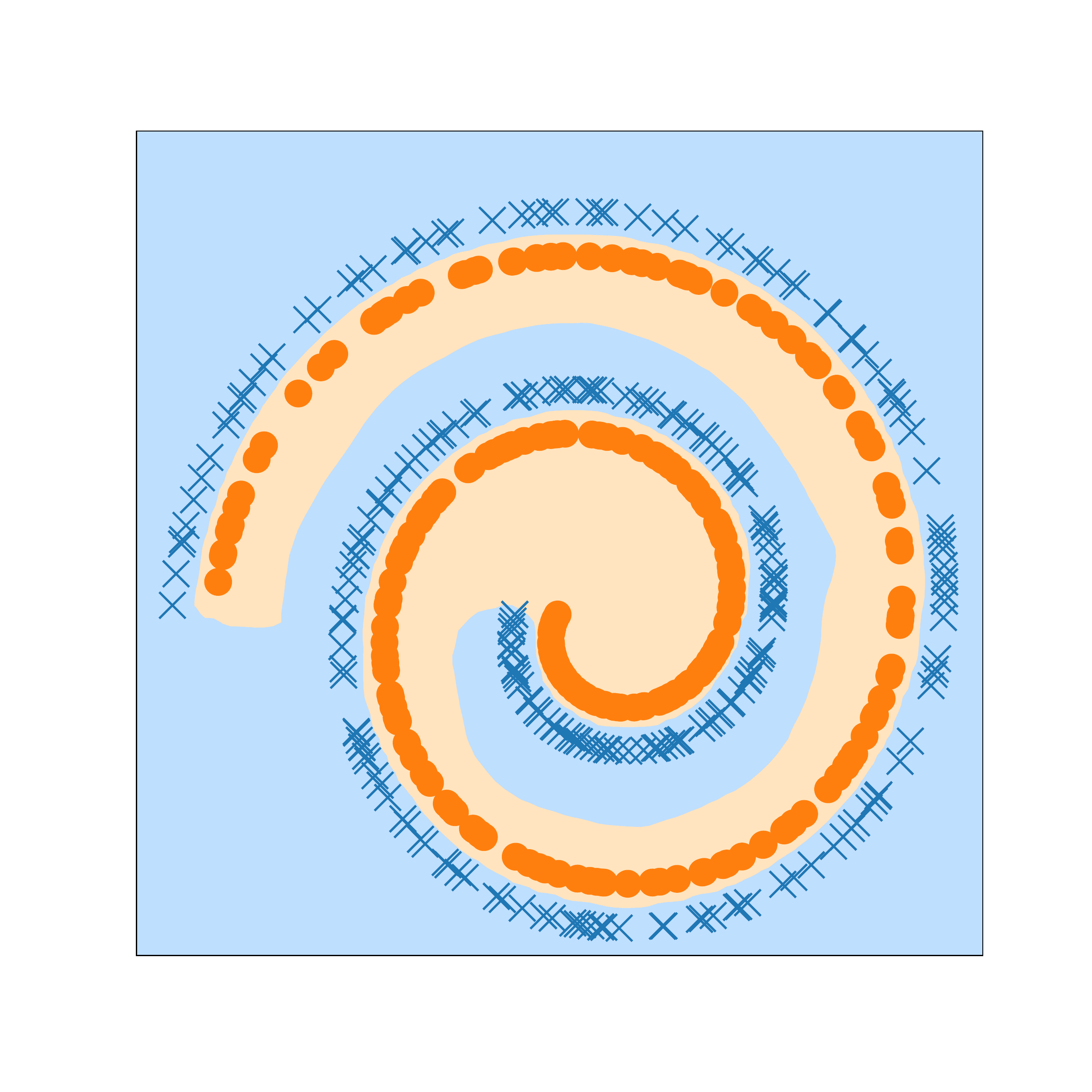}\vspace{0in} \vspace{-.1in}\\  \vspace{-.1in}
(a) Continuous Swiss Roll &  (b) GPM & (c) GPM+K & (d) Multi Task \\
\end{tabular}
\caption{The Continual Swiss Roll dataset consisting of 50 binary classification tasks (a) and the final decision boundaries of different methods, after having learned the 50 tasks sequentially (b)-(d). GPM+K denotes GPM with k-winner activations. %Quantitative and qualitative results show that GPM+K performs the best among different variations of GPM. 
We note that contrary to permuted-MNIST, and due to the high level of similarity between consecutive tasks, adding heterogeneous dropout does not improve the performance.}
\label{fig:all_swiss}
\end{figure}

\begin{figure}[t!]
    \centering
    \includegraphics[width=.9\linewidth]{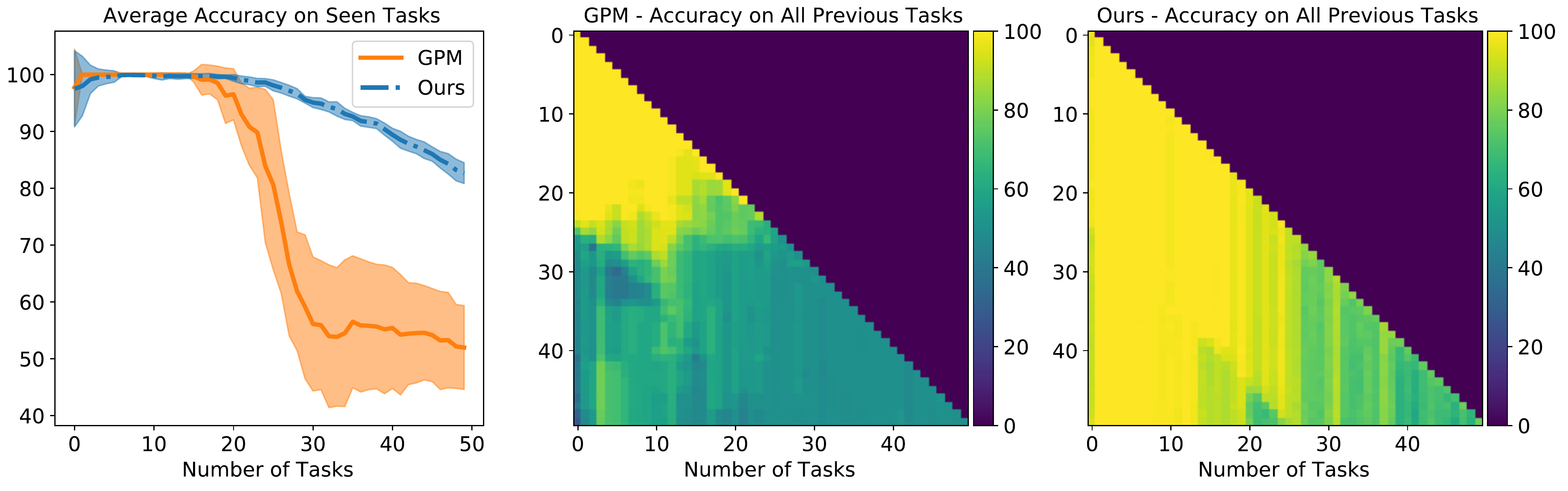}
    \caption{Performance analysis of GPM in comparison with our proposed method on the Continual Swiss Roll dataset with 50 tasks (as seen in Figure \ref{fig:all_swiss}). The results are averaged over 10 runs.}
    \label{fig:Swiss2}
\end{figure}

In this section, we first introduce \emph{Continual Swiss Roll} as a lightweight and easily interpretable, yet challenging, synthetic benchmark for continual learning. Continual Swiss Roll is generated from two classes of two-dimensional Swiss rolls, where the dataset is shattered into $T$ binary classification tasks according to their angular positions on the roll (See Figure \ref{fig:all_swiss} (a)). The continual learner needs to solve the overall Swiss roll binary classification problem by only observing the sequence of tasks. In addition to being simple and interpretable, one can arbitrarily increase the number of tasks and there is also an inherent notion of similarity between tasks in our proposed Continual Swiss Roll (according to their angular location). 

Here, we consider a Continual Swiss Roll problem with $T=50$ binary classification tasks. We use two single-head multi-layer perceptrons with two hidden layers of size $1024$, one using Rectified-Linear Unit (ReLU) activations (i.e., Dense) and the other using the k-winner activations with $k=64$. For this problem, we set $\alpha=0$, as the consecutive tasks share much similarities, and we would like to get forward transfer. We solve the continual learning using GPM. Figure \ref{fig:all_swiss} shows the learned decision boundaries after learning the 50'th task for GPM, and for our proposed method (i.e., GPM+k-winner). We see that adding sparse activations lead to a significant boost in performance and a much better preservation of the decision boundary in long sequences of tasks. We repeat this experiment $10$ times and report the average and standard deviation of the test accuracy over all seen tasks as a function of number of tasks in Figure \ref{fig:Swiss2}. In addition, in Figure \ref{fig:Swiss2}, we show the test accuracy of the current task as well as all previous tasks after learning each new task, as a lower triangular $50\times 50$ matrix for both methods. The $(t_1,t_2)$ element of this lower-triangular matrix contain the test accuracy of the model on Task $t_1$ immediately after training it on Task $t_2$. We see that the network with k-winner activations not only preserves the performance on old tasks better, but also leads to learning new tasks better. Figure \ref{fig:all_swiss} shows the final  decision boundaries (after learning the 50th task) learned with GPM, GPM+k-winner activations (ours), and the multitask learner (i.e., joint training).

\subsection{Permuted-MNIST}

\begin{table}[t!]
\small
\centering
\begin{minipage}[t]{.55\textwidth}
    \begin{tabular}{l|c c}
    & \multicolumn{2}{c}{Perm-MNIST (20 Tasks)}  \\
    & Accuracy & BT \\
    \hline
    EWC \citep{kirkpatrick2017overcoming} & 64.53 & -24.70\\
    ER \citep{rolnick2018experience} & 78.63 & -21.81 \\
    GPM \citep{saha2020gradient} & 81.23 & -14.70 \\
    AGEM \citep{chaudhry2018efficient} & 85.54 & -14.33 \\
    GEM \citep{lopez2017gradient} &{\bf 91.28} & {\bf -8.07} \\
    \hline 
    {\bf GPM + K + HD (Ours)} & {\bf 89.74} & {\bf -1.62} \\
    \hline 
    \end{tabular}
    \label{tab:my_label}
\end{minipage}%
\begin{minipage}[t]{.45\textwidth}
\raggedleft
    \begin{tabular}{l|c c}
    & \multicolumn{2}{c}{Perm-MNIST (50 Tasks)}  \\
    & Accuracy & BT \\
    \hline
    GPM & 36.56 $\pm$ 0.69 & -54.28 $\pm$ 0.75\\
    GPM + K  & 54.96 $\pm$ 1.92 & -31.03 $\pm$ 1.76 \\
    GPM + RD  & 33.60 $\pm$ 1.40 & -50.85 $\pm$ 1.63 \\
    GPM + HD  & 28.30 $\pm$ 2.02 & -20.55 $\pm$ 2.27 \\
    GPM + K + RD & 52.90 $\pm$ 2.48 & -27.52 $\pm$ 2.15 \\
    \hline 
    {\bf GPM + K + HD} & {\bf 61.09 $\pm$ 0.87} & {\bf -21.57 $\pm$ 0.39} \\
    \hline 
    \end{tabular}
\end{minipage}
    \caption{Final accuracy and backward transfer on Permuted-MNIST with 20 tasks, for benchmark approaches vs our proposed approach (left). Ablation study on 50 tasks of Permuted-MNIST, averaged over 5 runs. Abbreviations correspond to: K (k-winner), RD (Random Dropout), and HD (Heterogeneous Dropout). Our method (GPM + K + HD) achieves the highest accuracy and comparable backward transfer to other combinations of modules (right)}
    \label{tab:ablation_table}
\end{table}

Here, we perform our experiments on the Permuted-MNIST continual learning benchmark with $50$ tasks. Each of the 50 Permuted-MNIST tasks is
a 10-classes classification problem obtained from a random permutation (fixed for all images of a task) of the MNIST dataset, where the pixels
in each figure are permuted according to the permutation rule. For our models, similar to the models used in Section \ref{sec:lowDim}, we use multi-layer perceptron with two hidden layers of size $512$, and with ReLU activations (i.e., dense resulting in GPM), and with k-winner activations for $k=8$ and $32$. For dropout, we use $\alpha=24/k$ (i.e., $3$ for $k=8$ and $0.75$ for $k=32$). 

\begin{figure}[t!]
    \centering
    \includegraphics[width=0.5\linewidth]{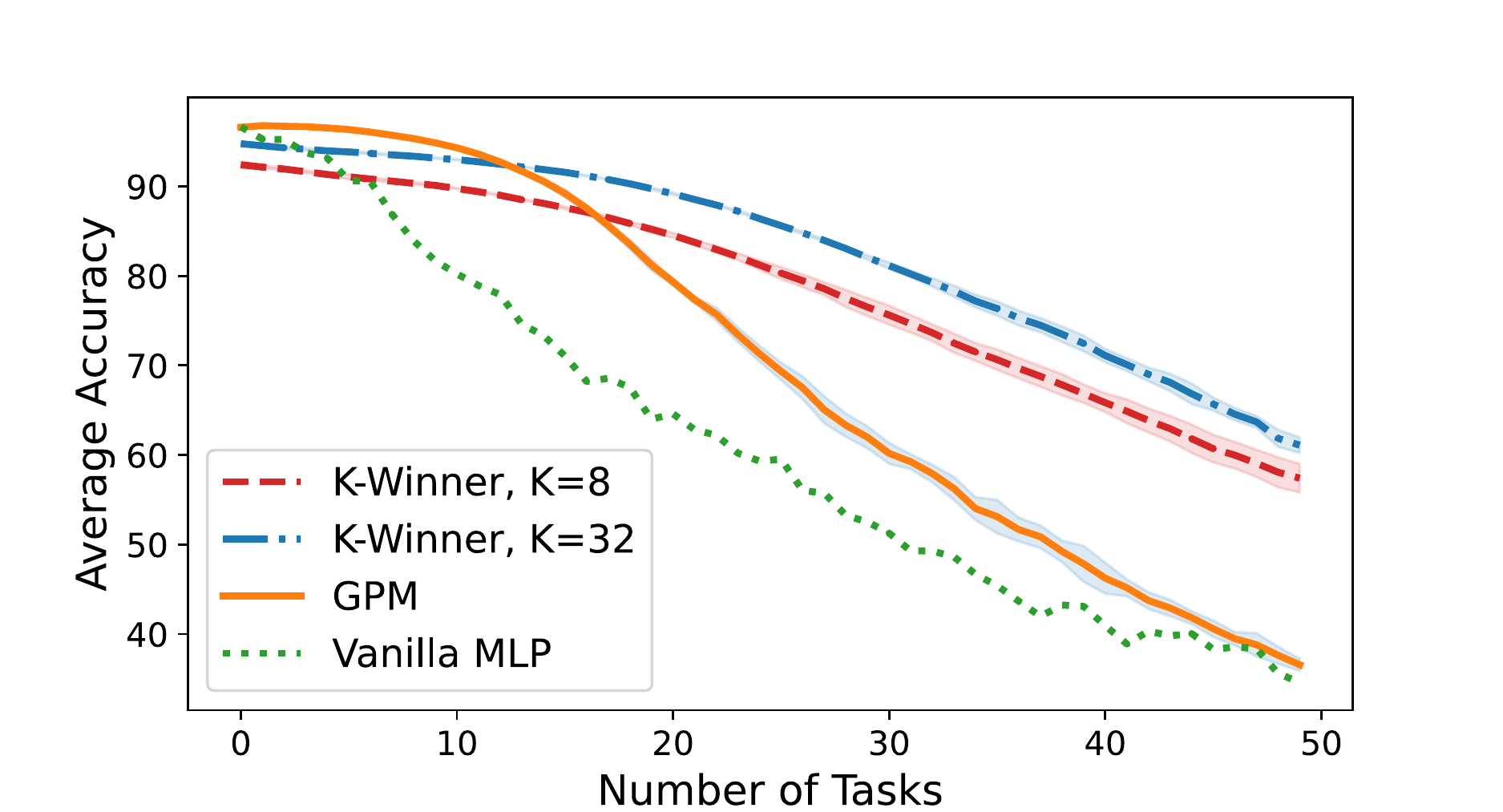}\hfill
    \includegraphics[width=0.5\linewidth]{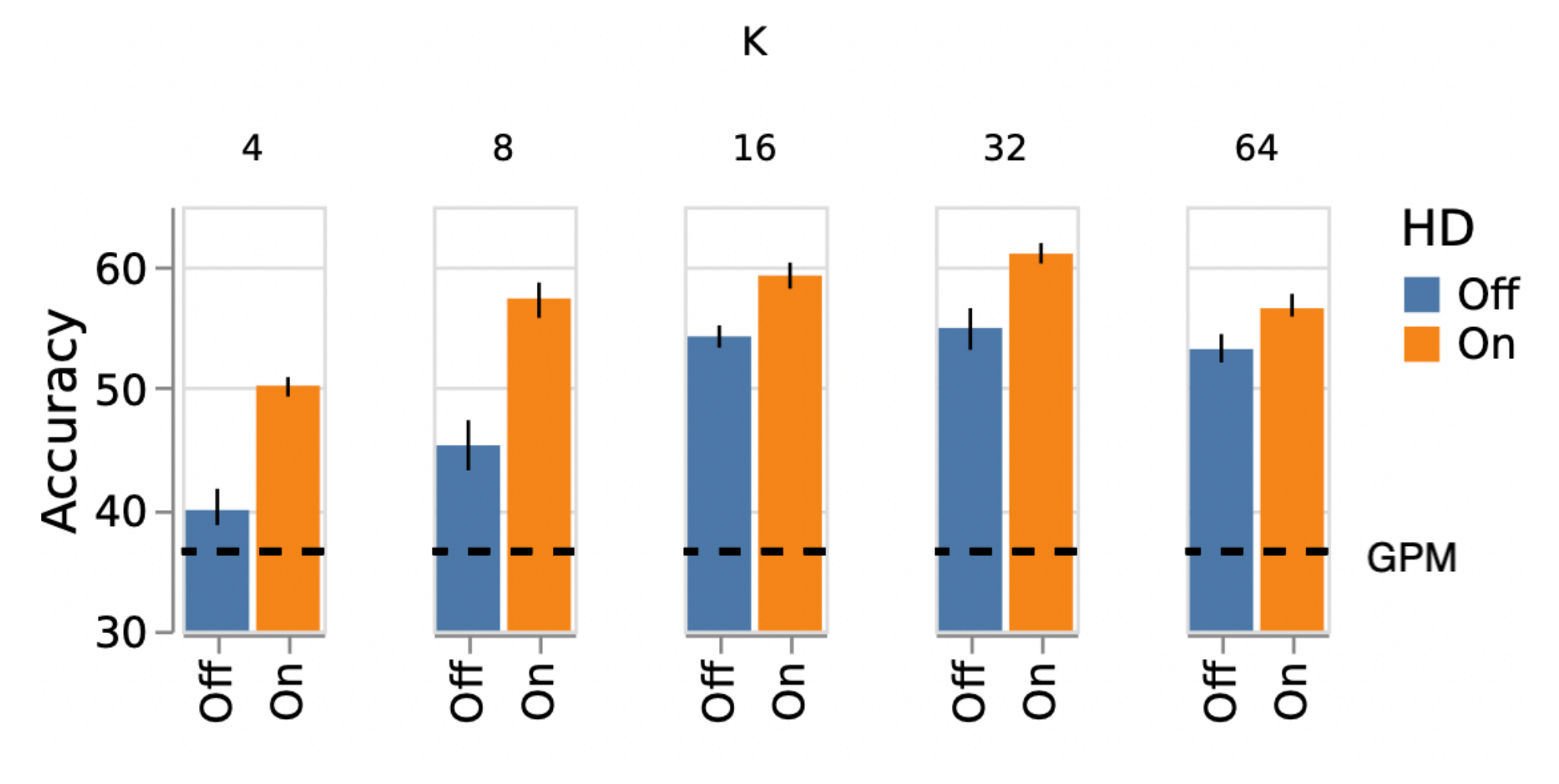}
    \caption{The results of our proposed approach with k-winner sparsity and heterogeneous dropout on the Permuted-MNIST continual learning benchmark with $50$ tasks (a). An ablation study quantifying the effect of sparsity and heterogeneous dropout in learning 50 Permuted-MNIST tasks. We see that sparsity and our heterogeneous dropout contribute to the performance boost in a complementary manner.}
    \label{fig:perm}
\end{figure}

Figure \ref{fig:perm} (left) shows the average accuracy of the models on previously seen tasks, as a function of the number of tasks. The results are aggregated from 5 runs. We can see that our proposed networks with k-winner sparsity and heterogeneous dropout significantly out-perform the GPM algorithm applied to a dense model. In addition to GPM, we compared our method to some of the benchmark algorithms in the literature including Elastic Weight Consolidation \citep{kirkpatrick2017overcoming}, Experience Replay \citep{rolnick2018experience}, Gradient Episodic Memory \citep{Lopez-Paz:2017}, and Averaged Gradient Episodic Memory \citep{chaudhry2018efficient} on 20 Task Permuted-MNIST. Table \ref{tab:my_label} shows the results of the benchmark algorithms compared to ours. We generate the results of other methods using the wonderful code repository provided by \cite{lomonaco2021avalanche,lin2021the}.

Lastly, we performed various ablation studies to shed light on the effect of each proposed component.  First, we look at the effect of $k$ in the k-winner activations on our 50-tasks permuted-MNIST experiment with and without heterogeneous dropout and compare it with vanilla GPM. The results of this experiment are reported in Figure \ref{fig:perm} (right). This experiment demonstrates that both sparsity and heterogeneous dropout play critical roles in improving the performance of GPM when learning a long sequence of tasks. %We show that GPM benefits from a large performance boost when the underlying network is sparsely activated, and heterogeneous dropout on a sparsely activated network significantly increases the performance boost of GPM in learning from a large number of tasks (see Figure \ref{fig:perm}).
To further provide insights into the performance of the proposed components, we ran a full ablation study with $k=32$, and also compared our proposed heterogeneous dropout with random dropout. The results of our full ablation study for this problem is shown in Table \ref{tab:ablation_table}.

% \begin{figure}
%     \centering
%     \includegraphics[width=0.5 \linewidth]{plots/Ablation.pdf}
%     \caption{Caption}
%     \label{fig:my_label}
% \end{figure}

% \begin{figure}[t]
%     \centering
%     \includegraphics[width=0.5 \linewidth]{plots/acc-task.pdf}
%     \caption{Caption}
%     \label{fig:my_label}
% \end{figure}

\subsection{CIFAR-100}
\begin{table}[t!]
    \centering
    \begin{tabular}{l|c c | c c}
    & \multicolumn{2}{c|} {CIFAR-100 Super Class (5 Tasks)} & \multicolumn{2}{c}{SuperDog-40 (8 Tasks)}\\
     & Accuracy & BT & Accuracy & BT \\
    \hline 
     GPM &  28.84 $\pm$ 0.64  & -22.51 $\pm$ 1.06 & 39.66 $\pm$ 1.06 &  -29.88 $\pm$ 1.30\\
    GPM + K & 26.57 $\pm$ 0.74  & -18.60 $\pm$ 0.95 & 37.91 $\pm$ 1.79 & -25.73 $\pm$ 1.69 \\
    GPM + RD & 24.86 $\pm$ 0.51  & -23.34 $\pm$ 0.74 & 38.72 $\pm$ 2.33 & -32.61 $\pm$ 3.11 \\
    GPM + K + RD & 17.55 $\pm$ 0.68 & -11.87 $\pm$ 0.68 & 36.73 $\pm$ 1.51 & -24.28 $\pm$ 2.19 \\
    \hline
    GPM + K + HD &  29.77 $\pm$ 0.41 & {\bf -9.55 $\pm$ 0.38} & 42.58 $\pm$ 0.76 & {\bf -23.17 $\pm$ 1.58} \\
    GPM + HD & {\bf 32.11 $\pm$ 0.71}  & -11.09 $\pm$ 0.79 & {\bf 42.82 $\pm$ 0.96} & -26.60 $\pm$ 2.05  \\
    
    \hline
    \end{tabular}
    \caption{Averaged final accuracy and backward transfer (over 5 runs) of an experiment on five 20-superclass classification tasks on CIFAR-100 (middle) and eight 5-superclass classification tasks on SuperDog-40 (right). Both experiments use a single-head AlexNet as their backbone. Since AlexNet has random dropout in its architecture, GPM and GPM + RD correspond to AlexNet without dropout and regular AlexNet, respectively.}
    
    %Results show that adding our extended version of k-winner on top of the CNN layers, does not hurt or improve the performance in a multi-head setup. }
    \label{tab:cnn_results}
\end{table}

In this section, we extend our results on MLPs to Convolutional Neural Networks (CNNs), and demonstrate the practicality of the k-winner activations and Heterogeneous Dropout beyond MLP architectures. We first point out that many of the existing benchmarks in the continual learning literature, e.g., split CIFAR-10, or split CIFAR-100, contain few tasks and do not really reflect the scalability issues of the existing approaches with respect to the number of tasks. Moreover, the majority of the existing approaches focus on multi-head neural architectures, while the arguably more challenging `domain incremental learning' \cite{van2019three} remains comparably understudied. Following the work of \cite{ramasesh2021anatomy}, here we use the `distribution shift CIFAR-100' as a domain incremental continual learning setting with five tasks. Each task, in this setting, is a 20-class classification problem, which contains one subclass from each superclass. We used AlexNet \cite{krizhevsky2012imagenet} as our backbone. Table \ref{tab:cnn_results} shows our results for this experiment. We observe that AlexNet with Heterogeneous Dropout outperforms other networks. We also noticed that k-winner activations combined with Heterogeneous Dropout results in the least amount of forgetting at the cost of a slight drop in the accuracy. 

\subsection{ImageNet SuperDog-40}

Here, we introduce ImageNet SuperDog-40 as an effective continual learning benchmark for domain incremental learning. SuperDog-40 is a subset of ImageNet \cite{deng2009imagenet} with images from $40$ dog breed classes. These dog breeds belong to five superclasses, namely 1) sporting dog, 2) terrier, 3) hound, 4) toy dog, and 5) working dog, resulting in eight five-way classification tasks. Similar to the `domain shift CIFAR-100' experiment, here we use an AlexNet to learn these 8 tasks. The results are shown in Table \ref{tab:cnn_results}. Consistent with our results on CIFAR-100, here we observe that the models containing Heterogeneous Dropouts and k-winner activations are the two top performers.

\section{Conclusion}
This paper studied the effects of sparsity and non-overlapping neural representations in the Gradient projection memory (GPM) framework in continual learning. In addition to providing a power-efficient sparsely activated network, we showed that a network trained with k-winner activations has low-dimensional neural activation subspaces. The low-dimensionality of the neural activation subspaces translates to having a large null-space, which leads to lower gradient projection error for learning new tasks in GPM. 
%\hamed{COMMENT: we do not really emphasize that GPM assumes small eigen values are the nullspace so even the projected gradient may affect the old tasks. Should we discuss this better in the introduction?} 
% \soheil{COMMENT: I will add a few sentences in the Introduction.}
% \hamed{Great, thanks!}
Moreover, we proposed a heterogeneous dropout mechanism, which encourages non-overlapping patterns of neural activations between tasks. We showed that heterogeneous dropout complements the $k$-sparse activations of the k-winner framework and significantly improves performance when using the GPM algorithm to learn a long sequence of tasks. We introduced \emph{Continual Swiss Roll} and \emph{ImageNet SuperDog-40} as two benchmark problems for domain incremental learning. Continual Swiss Roll is a lightweight and interpretable yet challenging, continual learning synthetic benchmark, while ImageNet SuperDog-40 is a subset of ImageNet with hand-picked dog images from 40 different breeds that belong to 5 superclasses. Finally, we analyze our proposed approach on the Continual Swiss Roll, Permuted-MNIST, CIFAR-100, and ImageNet SuperDog-40 datasets and provide an ablation study to clarify the contribution of each proposed component.

\section{Acknowledgement}
SK and HP were supported by the Defense Advanced Research Projects Agency (DARPA) under Contract No. HR00112190135. VB was supported by DARPA under Contract No. HR00112190130.

\bibliography{sgpm}
\bibliographystyle{collas2022_conference}

% \appendix
% \section{Appendix}
% You may include other additional sections here.

\end{document}